\pdfoutput=1
\documentclass[11pt]{article}

\usepackage[preprint]{acl}

\usepackage{times}
\usepackage{latexsym}

\usepackage[T1]{fontenc}
\usepackage[utf8]{inputenc}

\usepackage{microtype}

\usepackage{inconsolata}

\usepackage{graphicx}

\usepackage{enumitem} % for customizing enumerate labels

\title{Spacewalker: Traversing Representation Spaces for Fast  Interactive Exploration and Annotation of Unstructured Data}

\author{Lukas Heine\textsuperscript{1,3}, Fabian Hörst\textsuperscript{1,3}, Jana Fragemann\textsuperscript{1}, Gijs Luijten\textsuperscript{1,4}, Jan Egger\textsuperscript{1,4}, Fin H. Bahnsen\textsuperscript{1},\\{\bf M. Saquib Sarfraz}\textsuperscript{2}, {\bf Jens Kleesiek}\textsuperscript{1,3,5} \and {\bf Constantin Seibold}\textsuperscript{1}\\
        \textsuperscript{1}Institute for AI in Medicine, University Hospital Essen\\
        \texttt{\{firstname.lastname\}@uk-essen.de}\\
        \textsuperscript{2}Mercedes-Benz Tech Innovation\\%\\\texttt{saquibsarfraz@gmail.com}
        \textsuperscript{3}Cancer Research Center Cologne Essen, West German Cancer Center Essen, University Hospital Essen\\
        \textsuperscript{4}Institute of Computer Graphics and Vision, Graz University of Technology\\
        \textsuperscript{5}Department of Physics, TU Dortmund University\\
        }

\begin{document}
\maketitle

\begin{abstract}
In industries such as healthcare, finance, and manufacturing, analysis of unstructured textual data presents significant challenges for analysis and decision making. Uncovering patterns within large-scale corpora and understanding their semantic impact is critical, but depends on domain experts or resource-intensive manual reviews. In response, we introduce Spacewalker in this system demonstration paper, an interactive tool designed to analyze, explore, and annotate data across multiple modalities. It allows users to extract data representations, visualize them in low-dimensional spaces and traverse large datasets either exploratory or by querying regions of interest. We evaluated Spacewalker through extensive experiments and annotation studies, assessing its efficacy in improving data integrity verification and annotation. We show that Spacewalker reduces time and effort compared to traditional methods.
    The code of this work is available \href{https://anonymous.4open.science/r/NAACL-78A4/}{here}.
\end{abstract}
\begin{figure*}
    \centering
    \includegraphics[width=0.93\linewidth]{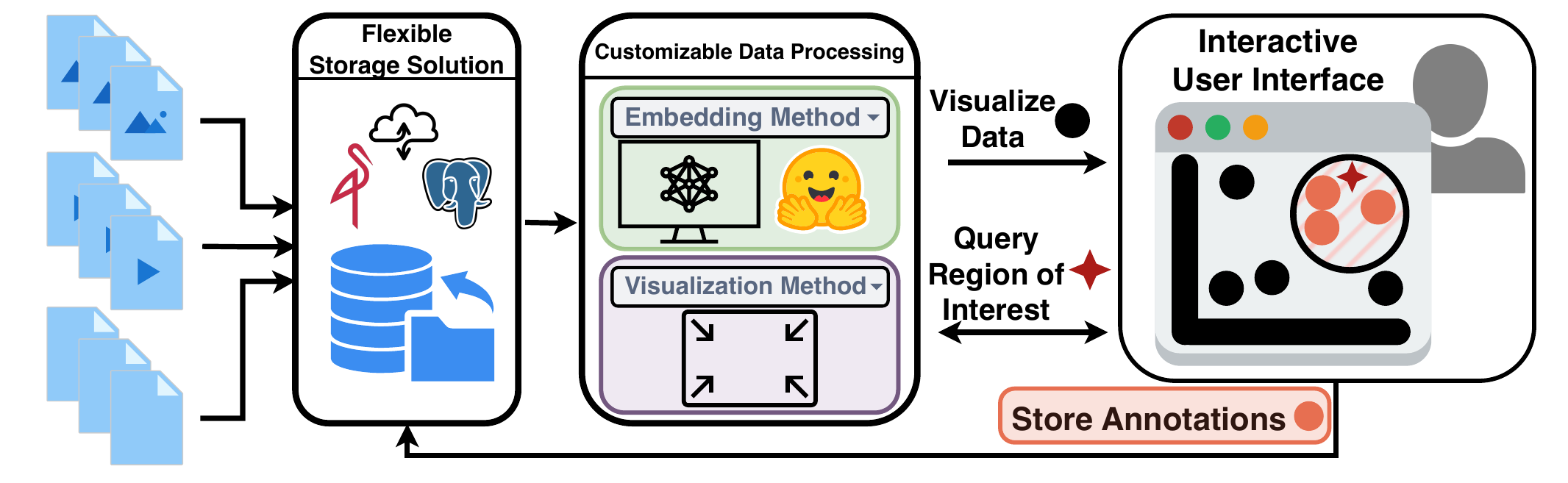}
    \caption{Spacewalker is a tool designed for pattern discovery within extensive multimodal datasets, employing arbitrary neural networks and visualization techniques. Once the data is visualized, users can dynamically examine the dataset through simple mouse interactions and multimodal queries to pinpoint samples.}
    \label{fig:teaser}
\end{figure*}

\section{Introduction}

Rapid expansion of data in industries such as healthcare or finance presents both challenges and opportunities for information retrieval and data-driven decision making \cite{rydning2018digitization}. Projections estimate that the global data volume will surpass 180 zettabytes \cite{petroc2023volume}, with up to 80\% being unstructured data \cite{edge2024unstructured}, such as text, images, and multimedia. 

The advent of LLM-based Retrieval-Augmented Generation (RAG) frameworks \cite{xing2024designing, Liu_LlamaIndex_2022} has provided powerful solutions to extract knowledge from unstructured datasets. However, they can be vulnerable to issues such as data contamination, where poisoned or low-quality samples can adversely influence downstream tasks \cite{zou2024poisonedrag}. Although manual inspection and machine learning models can assist in identifying such problematic data points, it often turns to finding the needle in a haystack. Similarly, for the categorization of large datasets, projects often rely on tedious manual annotation on a sample-bases via tools like LabelStudio. 

We argue that traversing datasets on a sample basis is an inefficient use of annotator resources and could be remedied through different visualization methods. 
Despite advances in visualization tools \cite{renumics, abadi2016tensorflow, paurat2013invis}, several challenges remain: Many tools struggle to integrate multiple data modalities (e.g. text, images, video), are not optimized for real-time interaction with large datasets, lack adaptability to evolving models, or offer limited support for customizable 2D and 3D visualizations, and in general require expert knowledge. Moreover, existing tools rarely integrate annotation and storage functionalities for managing discovered insights like tags or semantic associations. To address these limitations, we present Spacewalker, an interactive tool for the exploration, annotation, and analysis of unstructured datasets across diverse modalities, with a focus on text data. Spacewalker enables users to upload and process datasets, extract semantic representations using pre-trained or custom embedding models, visualize embeddings in low-dimensional spaces, and perform multimodal querying to explore semantic relationships or detect anomalies.

Through extensive user studies, we demonstrate that Spacewalker significantly accelerates data annotation and improves user interaction compared to conventional methods. For tasks involving the identification of corrupted or mislabeled datasets, its latent space visualization and multimodal querying system enable rapid and accurate detection of critical data points. Furthermore, the intuitive interface of the tool allows non-technical users to interact with complex datasets without specialized expertise.

% We release Spacewalker as an open-source project to promote accessibility and extensibility across various industries and research domains, addressing the growing need for interactive and user-friendly data exploration systems.

% We state our contributions as follows:
% Firstly, we provide an open-source interactive tool for multimodal data annotation as well as visualization, supporting diverse embedding models, dimensionality reduction methods (DRMs), and real-time 2D/3D exploration. Secondly, we provid ethe findings of a comprehensive user study evaluating Spacewalker’s accuracy, interaction efficiency, and semantic integrity. Thirdly, we show empirical recommendations for embedding and DRMs to produce effective data representations.
%\set

\section{Related Work}
\begin{figure*}[!t]
    \centering
    \includegraphics[width=0.99\linewidth]{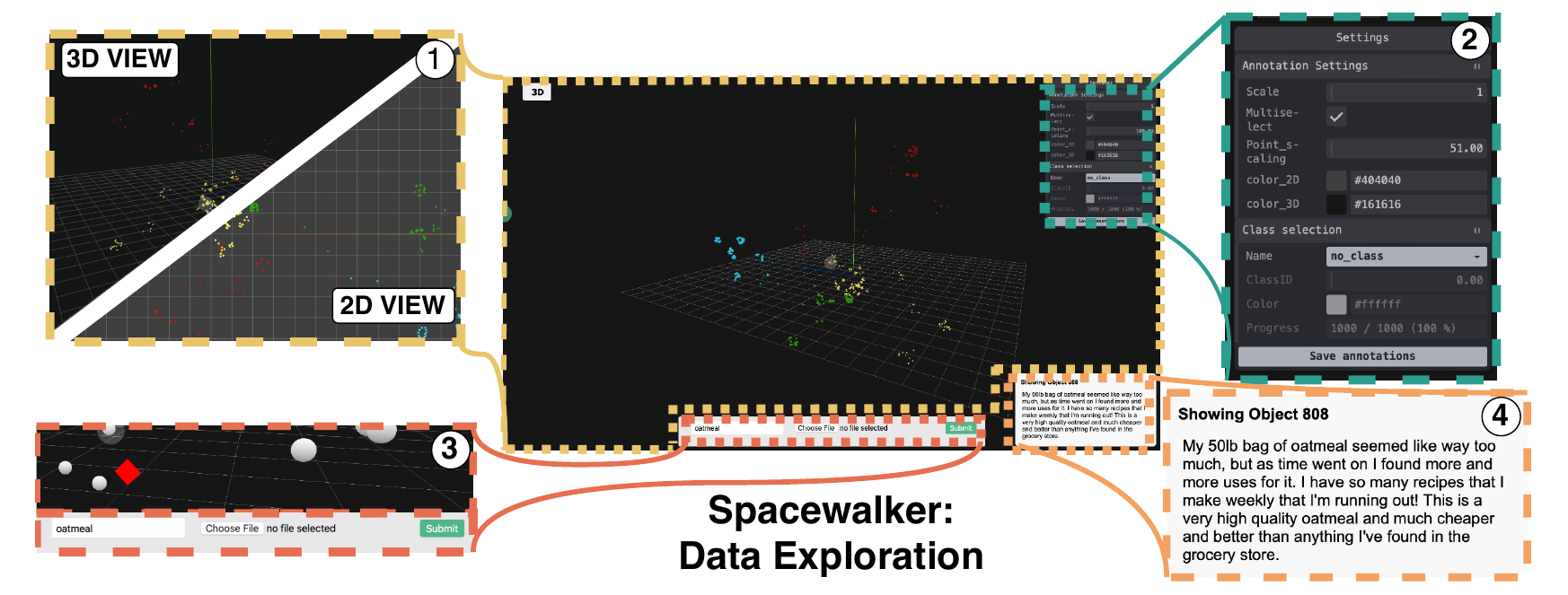}
    \caption{Main UI components of Spacewalker: Lower-dimensional representations (1), visualization and annotation parameters (2), query dialog (3) and data preview (4)}
    \label{fig:ui_components}
\end{figure*}
\noindent\textbf{Representations of unstructured data:} Representation learning is crucial for unstructured data analysis. In NLP, while models like Word2Vec \cite{mikolov2013efficient} and GloVe \cite{pennington2014glove} excel at capturing local semantics, Sent2Vec \cite{pagliardini2017unsupervised} addresses limitations in broader context understanding. Transformer models including BERT \cite{devlin2018bert} and GPT-4 \cite{achiam2023gpt} enhance document-level representations. In computer vision, CNNs \cite{lecun1989backpropagation, he2016deep} and ViTs \cite{dosovitskiy2020image} effectively learn image representations in supervised \cite{krizhevsky2012imagenet} or self-supervised settings \cite{zhang2022dino}. For video, earlier models \cite{tran2015learning,feichtenhofer2019slowfast} capture temporal dynamics, while transformers can \cite{arnab2021vivit,bertasius2021space} deal with long-range frame dependencies. Multimodal models link text and images \cite{radford2021learning,alayrac2022flamingo} or videos \cite{lei2021less}. These developments underscore the value of tools that exploit representations to enhance unstructured data interpretability. Spacewalker leverages these representations for seamless data exploration.

\noindent\textbf{Visualization of high-dimensional data:}
Dimensionality reduction methods (DRMs) simplify high-dimensional data for visualization, revealing patterns and relationships. Approaches such as stochastic neighborhood integration (SNE) \cite{hinton2002stochastic} and t-SNE \cite{van2008visualizing} prioritized local distance preservation, but are computationally expensive. UMAP \cite{mcinnes2018umap} improved scalability and preservation of global structure. More recent methods like h-NNE \cite{sarfraz2022hierarchical} introduce hierarchical clustering and real-time querying with minimal parameter tuning.
Although these techniques transformed data analysis, they often require significant programming expertise. Tools such as Scatter/Gather \cite{scatter}, Supervised PCA \cite{paurat2013supervised}, InVis \cite{paurat2013invis}, and PatchSorter \cite{walker2024patchsorter} address specific data types, such as text and histopathology images.
In contrast, Spacewalker integrates multimodal data, offering synchronized 2D and 3D visualizations and an extensible ecosystem of embedding methods. Its cross-domain querying allows navigation through text, image, and video inputs. By supporting diverse DRMs and embedding methods, Spacewalker enables interactive and customizable visualizations.

%\noindent\textbf{Exploration of Network Representations:}
%Data visualization plays a crucial role in debugging AI models, though its capacity for quantitative evaluation of deep learning performance is somewhat restricted \cite{balayn2023faulty}. Tools like TensorBoard \cite{abadi2016tensorflow} aid in visualizing embeddings, yet many researchers favor programmatic solutions, which can be complex and demand specific expertise \cite{balayn2023faulty}. Spacewalker addresses these challenges by delivering real-time, no-code insights into complex datasets and models. Equipped with features such as tagging, class associations, and direct storage capabilities, it offers an intuitive and flexible alternative for visualizing embeddings and diagnosing models. Its adaptability maintains its usefulness as machine learning technology evolves, making it a valuable tool for researchers.

\section{Spacewalker}

\noindent\textbf{Modalities:}
Spacewalker is designed to handle diverse data types such as text, images, or video. To achieve this, the tool stack incorporates an adaptable data storage solution, specifically an open-source S3 implementation. This enables advanced users to integrate it with current S3 storages or mirror data from other buckets.\\
\noindent\textbf{Data processing:}
Evaluating various combinations of embedding models and DRMs manually is labor-intensive. Spacewalker offers independent, reusable workflows for embedding strategies and DRMs, which can be effortlessly interchanged using drop-down menus. This significantly minimizes effort and enables users to visualize numerous combinations without any coding.\\
\noindent\textbf{Navigation:}
Static plots can significantly hinder data exploration. We support user-friendly mouse controls for zooming, panning, rotating, and annotating, enhancing interactivity beyond conventional static interfaces. We question the prevalence of 2D plots in data analysis by offering a novel, comprehensive 3D view. The added dimension effectively minimizes occlusion and enhances visibility of individual data points. In 3D, data selection involves determining depth via raycasting from the mouse origin. Orientation in these spaces can remain challenging, so a query dialog (Fig. \ref{fig:ui_components}.3) is provided to display the placement of text, image, or video queries within this spatial representation.\\
\noindent\textbf{Annotation:}
Regarding annotation, a primary advantage of Spacewalker is its use of pretrained models to identify similar samples by observing grouped data points. This capability allows for rapid and precise identification of similar points. Users can exploit the spatial configuration of the representation space by resizing the selector geometry to efficiently label large clusters of semantically similar points.\\
\noindent\textbf{Implementation Overview:} Spacewalker employs a microservice architecture with distinct containers assigned specific services. It utilizes PostgreSQL to store both 2D and 3D data points and metadata. MinIO S3 is used for storing unstructured data (embeddings, media), with previews generated by a webhook upon upload. This cloud storage system integrates projects without local server files. Model inference is carried out using the NVIDIA Triton server, establishing a REST API for model result retrieval. Django operates as the main webserver, directing microservice traffic and serving the user interface. For dimensionality reduction, Spacewalker supports all Scikit-learn workflows, as well as Scikit-learn-like implementations such as umap-learn, and openTSNE out of the box. The main graphical user interface (GUI) is rendered using three.js for an engaging and smooth user experience that supports complex computer graphics operations with GPU acceleration.
% \begin{figure*}[!t]
%     \centering
%     \includegraphics[width=1\linewidth]{figures/draw/overview.pdf}
%     \caption{Main UI components of Spacewalker: Lower-dimensional representation (yellow), settings (cyan), query dialog (red) and data preview (orange)}
%     \label{fig:ui_components}
% \end{figure*}

\section{User studies}

To test and validate the efficacy of our tool in practical settings, we recruited $n=20$ participants ($6$ female, $14$ male) aged 24 to 50 years with diverse IT backgrounds using snowball sampling. %Initially, three participants joined a preliminary study in which we tested a prototype. User feedback collected during the preliminary study indicated that several features were crucial to improving the user experience, which were subsequently implemented: A "CTRL + Z" undo mechanism to revert changes, a progress overview to display annotated and remaining samples, and a "paintbrush mode" allowing users to label points by dragging the right mouse button. Moreover, the color of the selector geometry was changed to always show the currently selected class, improving visibility. In addition, we opted to scale the coordinate system to reduce occlusion in clusters. 
We systematically evaluated Spacewalker's improvements in exploratory data analysis (EDA), data annotation, and data integrity verification. Firstly, a study was conducted to assess the influence of various models and DRMs (Section \ref{sec:dim_red}): Participants identified corrupted samples that should be excluded from the dataset, aiding in discerning preferred Spacewalker applications. Following this, an extensive study involved users annotating text and image datasets, assessing Spacewalker's efficiency across diverse annotation tasks, and confirming its practical robustness (Section \ref{sec:annotation}).

\subsection{Data processing}
\label{sec:dim_red}

This user study explores optimal combinations of embedding methods and DRMs to detect corrupted data sets, where corruption is defined as the existence of samples from classes not originally present. A randomized procedure determined whether participants viewed a "clean" or "corrupted" data set. To simulate corruption, between three and five samples from an external supercategory were added to the ImageNette dataset. We examined DRMs and embeddings in distinct settings. Participants in the "embedding" group were exposed to various configurations (CLIP + h-NNE, DinoV2 + h-NNE, and ResNet50 + h-NNE) presented in random order.
%We display the potential differences in visualization setup in Fig.\ref{fig:emb_comparisons}.
They were briefed on potential outliers (e.g., foreign food items in an ImageNet subset), allotted five minutes to evaluate the dataset, and recorded their observations in a questionnaire.

\begin{figure}[t]
    \centering
    \includegraphics[width=\linewidth]{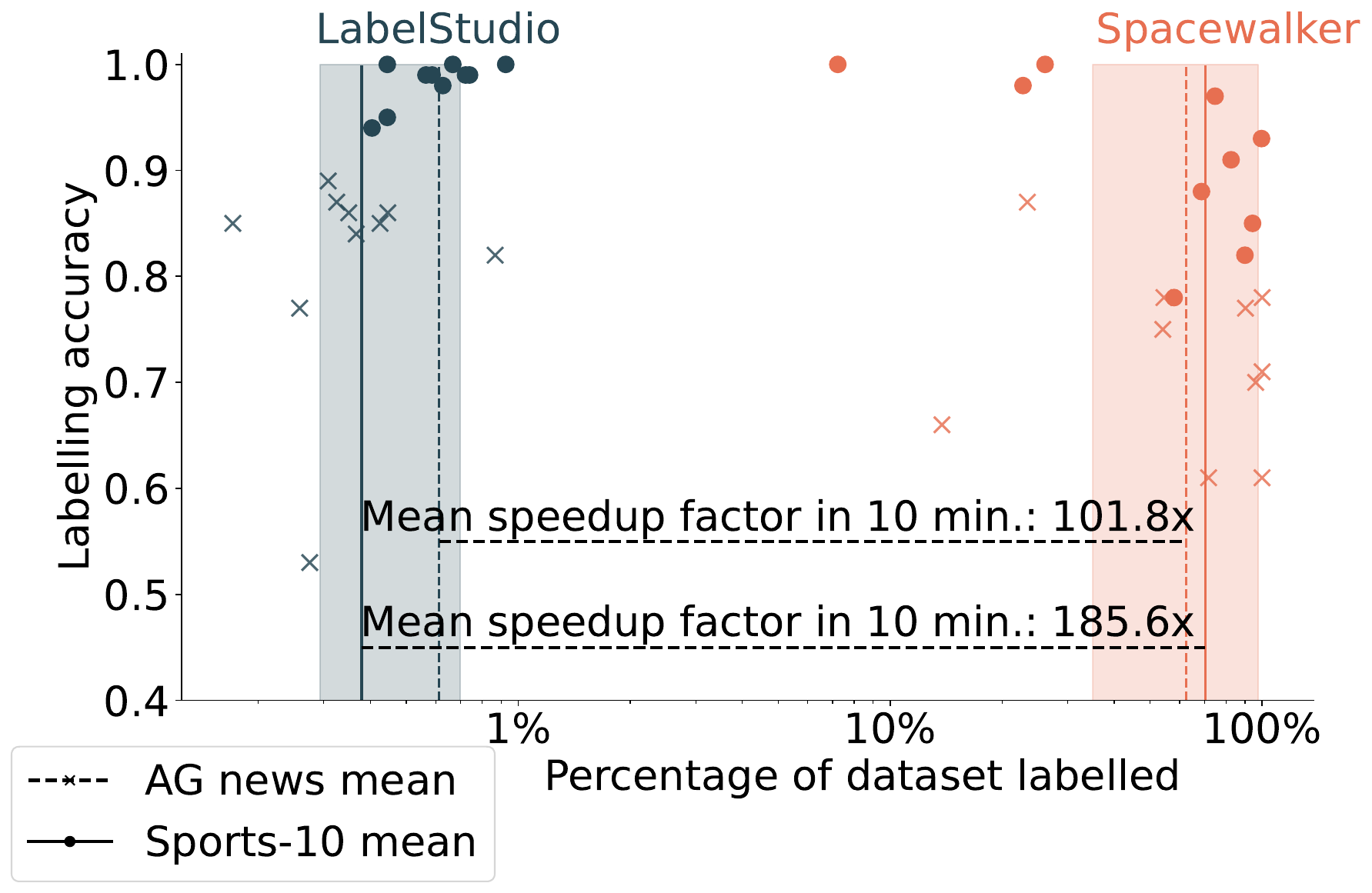}
    \caption{Labeling performances in LabelStudio (blue) and Spacewalker (orange) for text and image}
    \label{fig:annotation_comparison}
\end{figure}

\begin{figure}[!b]
    \centering
    \includegraphics[width=\linewidth]{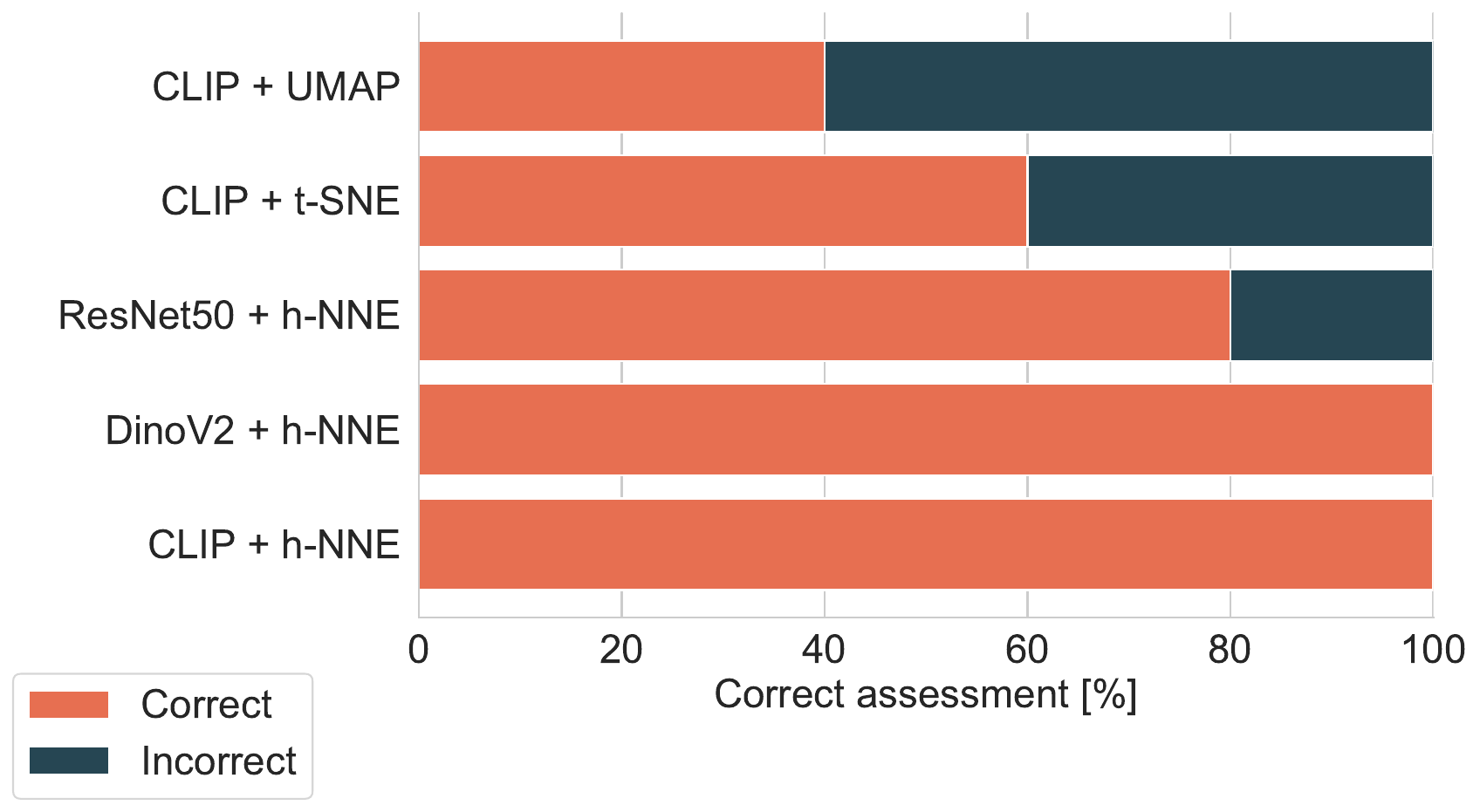}
    \caption{Participants' performances in the data integrity assessment user study for different combinations of encoders and DRMs}
    \label{fig:dia_results}
\end{figure}

\subsection{Annotation Process Assessment}\label{sec:annotation}
This study evaluates the annotation process in Spacewalker for text and image classification, using the AG News dataset \cite{zhang2015character} for text and the Sports-10 dataset \cite{trivedi2021contrastive} for images. LabelStudio \cite{tkachenko2020label}, a widely used annotation tool, was included as a baseline comparison. 20 participants were randomly assigned to either the "text" or "image" group. Participants received a brief tutorial and had unlimited time to familiarize themselves with their first randomly assigned tool, accompanied by a printed guide. They then annotated the dataset for ten minutes before completing a set of questionnaires, including the System Usability Scale (SUS) \cite{brooke1996sus}, NASA Task Load Index (NASA-TLX) \cite{hart1986nasa}, and free text comments. This process was repeated with the second tool. Observations of participant behavior and comments were recorded by study supervisors.

% \begin{figure*}[htbp]
%     \centering
%     \begin{tabular}{ll}
%         \includegraphics[width=.45\linewidth]{images/dr_ranks.png} & 
%         \includegraphics[width=.45\linewidth]{images/dia.png} \\
%         (a) User rankings.  & 
%         (b) Assessment Accuracy \\
% Top: embeddings, Bottom: dimensionality reduction &
        
%     \end{tabular}
%     \caption{User performances for correctly identifying samples that did not belong to the original dataset}
%     \label{fig:ranks}
% \end{figure*}

\begin{figure}[!bp]
    \centering
    \includegraphics[width=\linewidth]{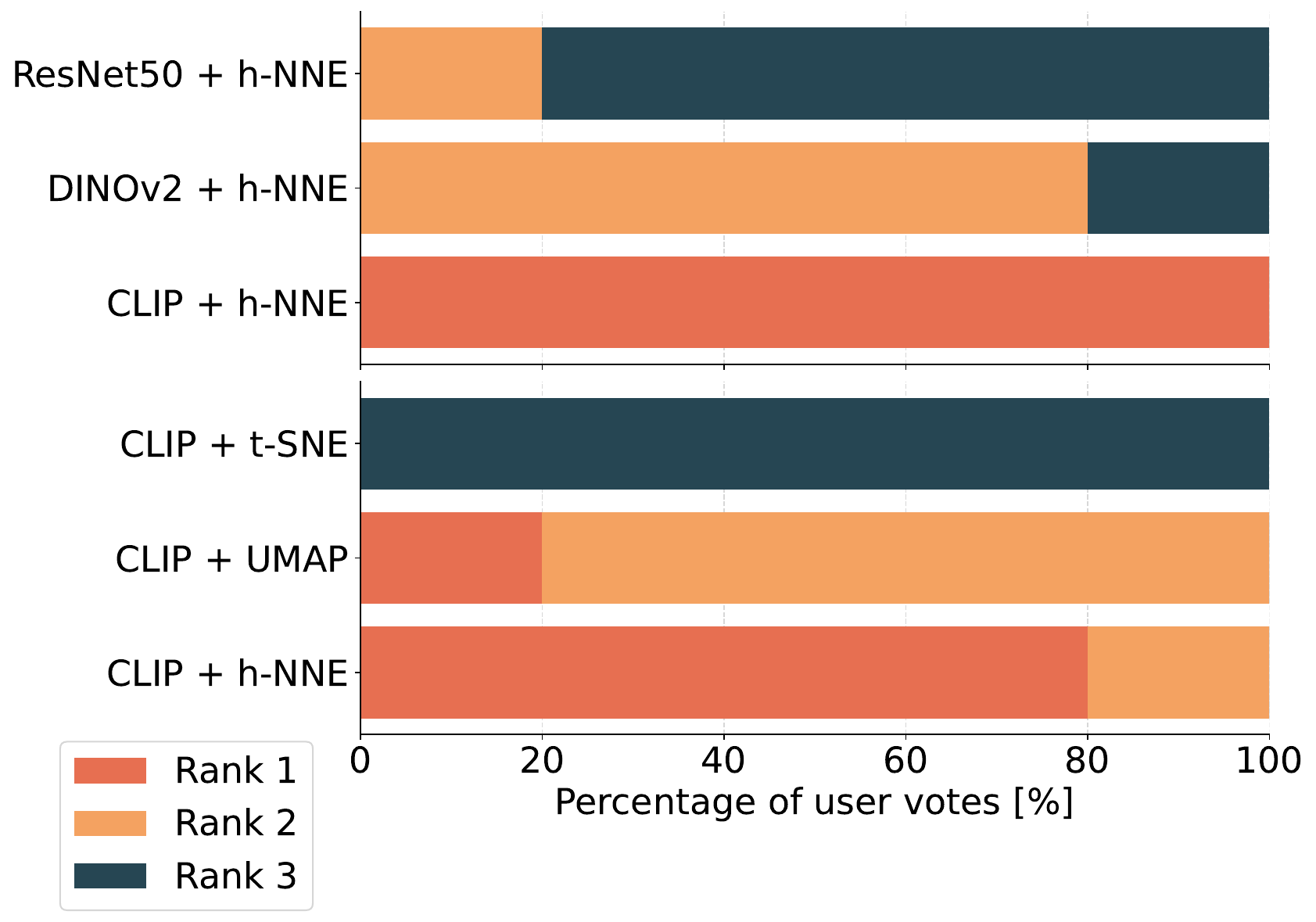}
    \caption{Participants' rankings of encoder and DRM combinations}
    \label{fig:user_ranks}
\end{figure}

\section{Results}
\label{sec:results}
\begin{figure*}[!t]
    \centering
    \includegraphics[width=0.95\linewidth]{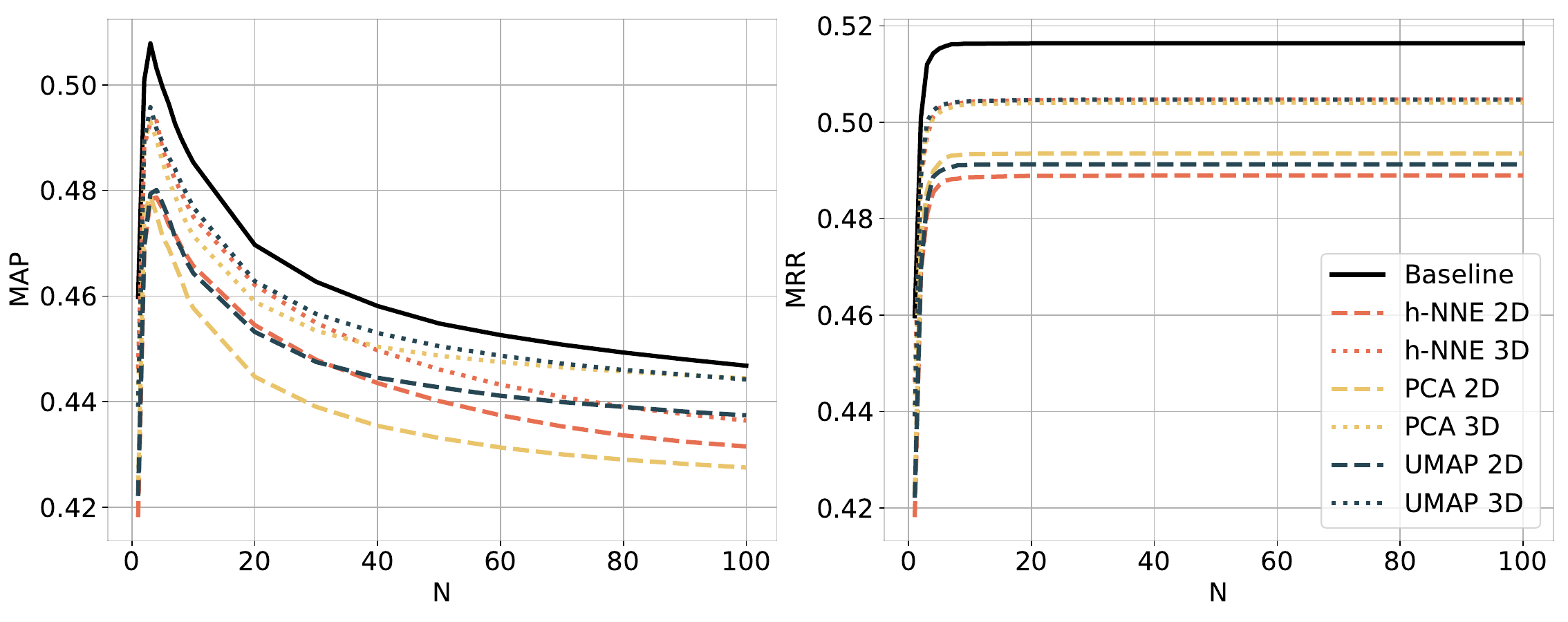}
    \caption{Mean Average Precision (MAP) (left) and Mean Reciprocal Rank (MRR) (right), adjusted for label frequency, for different methods on the OLID dataset}
    \label{fig:map}
\end{figure*}

We argue that it is possible to generate meaningful, low-dimensional representations of data from high-dimensional embeddings while preserving local neighborhoods and similar samples. To ground this hypothesis, we embed the train split of the OLID dataset \cite{zampierietal2019} using the BGE-M3 embedding model \cite{bge_m3} and calculate the label frequency adjusted Mean Average Precision (MAP) and Mean Reciprocal Rank (MRR) for the test split using the full, high-dimensional embeddings (baseline) as well as for various DRMs, preserving either two or three dimensions for visualizations (cf. Fig. \ref{fig:map}). 
Both metrics indicate that sophisticated DRMs are able to preserve relevant connections between samples effectively, as shown by only minor decreases in MAP and MRR. Intuitively, preserving three dimensions yielded marginally better results, hence our decision to support visualizations in 3D.

Selecting optimal combinations of embedding and DRM is a complex task due to the wide range of available options and modalities. To guide our recommendations, we integrate both objective performance assessments and subjective user ratings. In a first study, participants ranked the embeddings of CLIP, DinoV2, and ResNet50. While all models demonstrate viability when paired with h-NNE, CLIP and DinoV2 slightly outperform ResNet50 in terms of performance, as seen in Fig. \ref{fig:user_ranks}. User rankings, as illustrated in Fig. \ref{fig:user_ranks}, indicate a preference for CLIP. Notably, two participants provided the following feedback: \textit{"The ability to search by text was a nice-to-have."} (P5) and \textit{"I think text querying is more efficient than image search as you don't have to get images first."} (P4), underscoring the advantages of multimodal models that integrate text and image queries.
In a second study, DRMs were evaluated. Among them, h-NNE emerged as the most effective and was the preferred choice by a significant margin, as seen in Fig.\ref{fig:user_ranks}. Users tended to make the fewest errors in identifying corrupted datasets with h-NNE, followed by t-SNE and then UMAP. In terms of user ratings, the lower ranking of t-SNE may be attributed to its limitation in projecting new points. The participants particularly valued the responsiveness of h-NNE, with one noting: \textit{"I feel like queries in this setting (referring to h-NNE) are faster than the other (referring to UMAP), which makes interaction a little bit smoother"} (P2). Additional positive feedback on the search functionalities included: \textit{"Image and text querying are a nice feature."} (P1) and \textit{"I liked the search functionalities (images and text). The point scaling was essential; for this task, I preferred the 2D view. [...]"} (P2).
%This feedback highlights the utility of effective search functionalities and responsive dimensionality reduction in facilitating unstructured data analysis tasks.

% \begin{figure}
%     \centering
%     \includegraphics[width=1\linewidth, height=0.55\linewidth]{images/study.png}
%     \caption{Results of the System Usability Score (SUS) questionnaires (\textbf{left}) and NASA Task Load Index (NASA-TLX) questionnaires (\textbf{right})}
%     \label{fig:sus_and_nasatlx}
% \end{figure}

\begin{figure*}[!tp]
    \centering
    \begin{tabular}{cc}
         \includegraphics[width=0.5\linewidth]{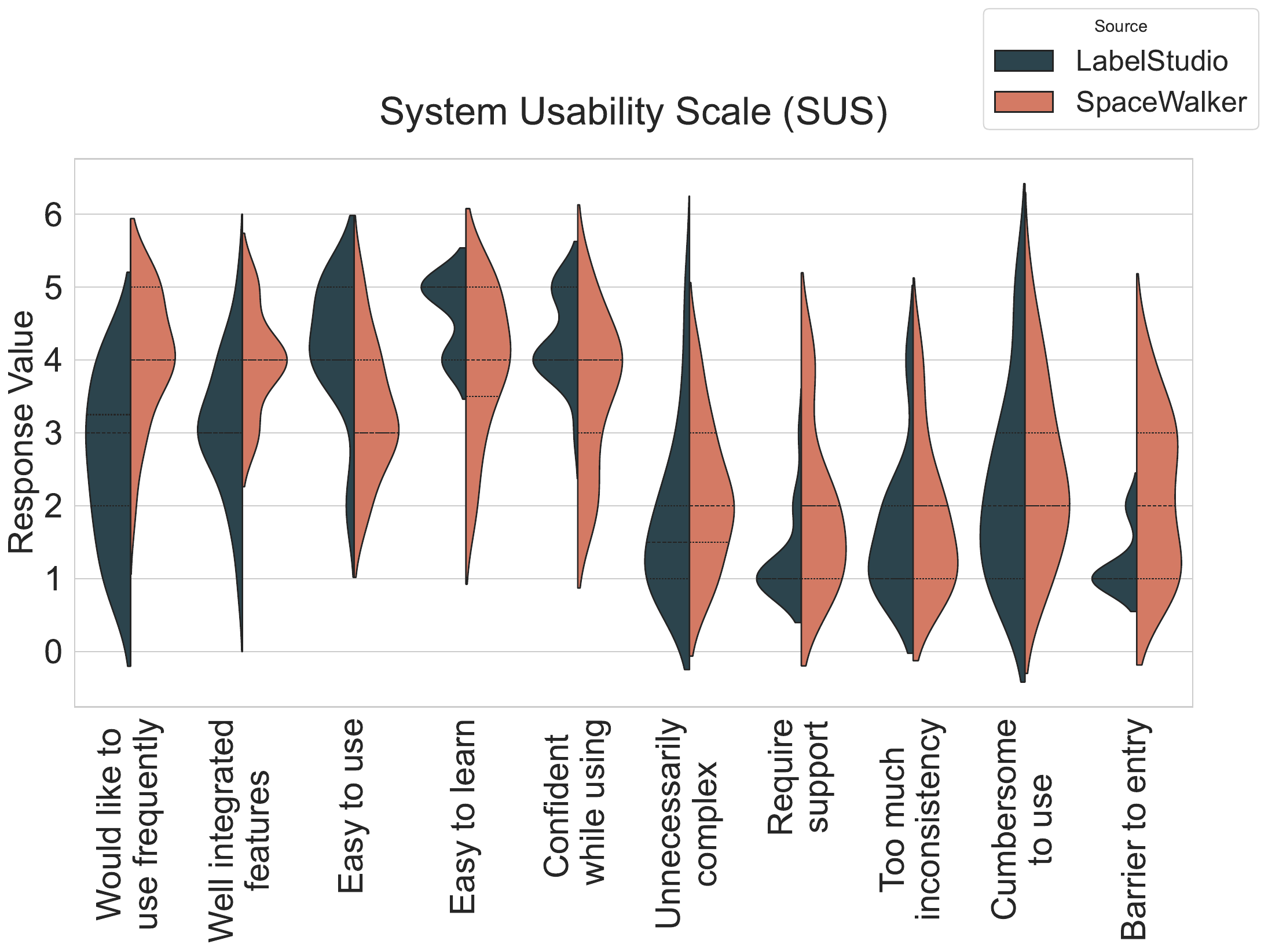} &  \includegraphics[width=0.5\linewidth]{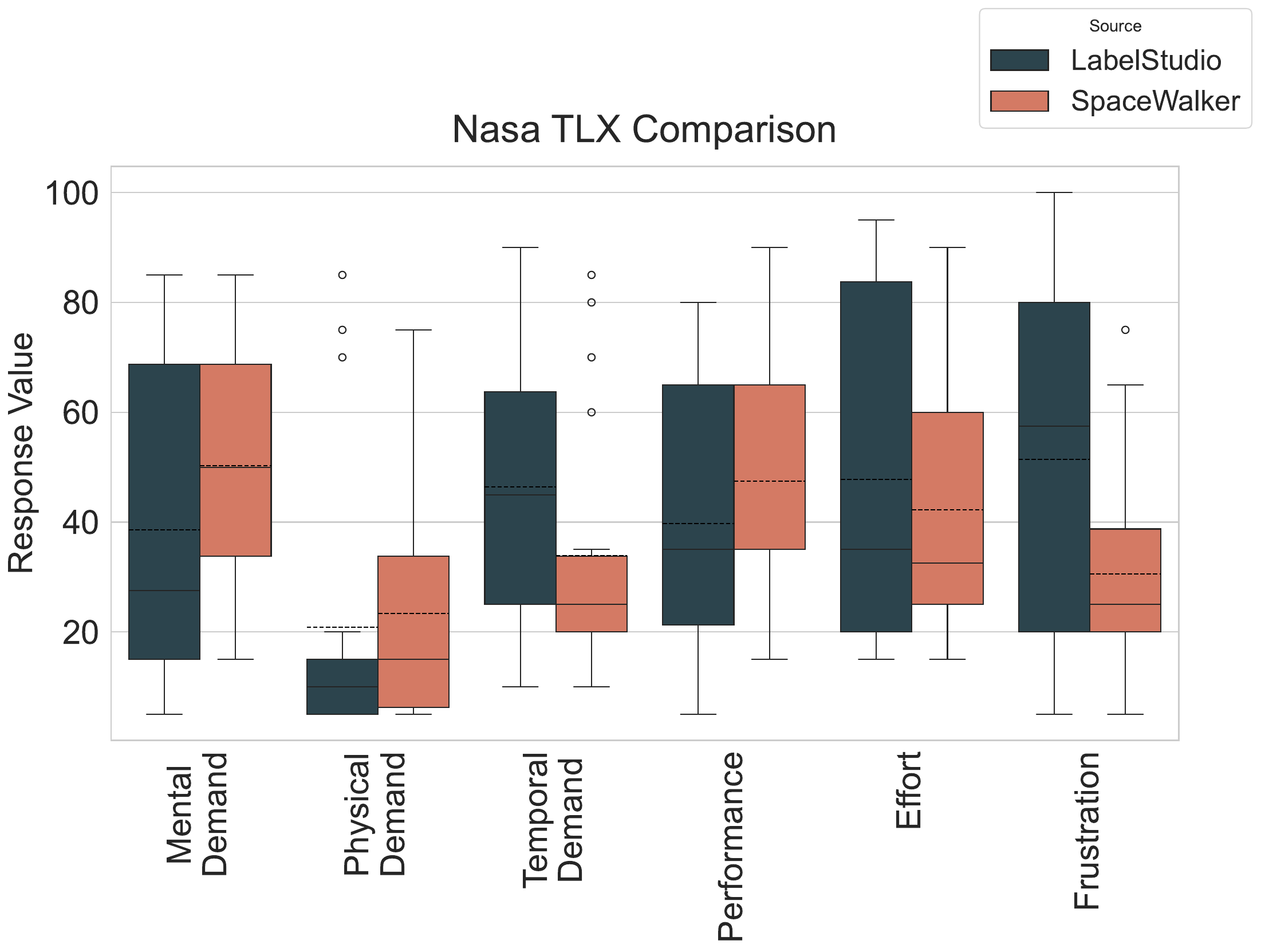}
    \end{tabular}
    \caption{Results of the System Usability Scale (SUS) questionnaire (left) and results of the NASA Task Loading Index (NASA-TLX) questionnaire (right)}
    \label{fig:nasatlx_sus}
\end{figure*}

\subsection{Data Annotation Performances}\label{sec:annotation_results}
%The following section evaluates the objective annotation performances both in terms of the number of samples labelled as well as as their accuracy in doing so. For a full, detailed overview over the performances of each individual participant, please refer to the appendix (\ref{sec:annotation_details}).
 To provide a complete picture, it is essential to consider the dataset complexity, as datasets with well-separable features are generally easier to annotate. To objectively measure the homogeneity of the dataset, we calculate the Normalized Mutual Information (NMI) between the K-Means clustering \cite{macqueen1967some} of the raw embeddings and the ground truth. The NMI for the Sports-10 dataset (images) was 0.94, indicating easier separability compared to AG News (text) with an NMI of 0.56.
Figure \ref{fig:annotation_comparison} shows annotation accuracy, with dots representing individual user performances for Spacewalker and LabelStudio. The shaded areas show the standard deviation. In the image task, participants labelled 168.1 samples on average with LabelStudio and 17,119.7 with Spacewalker in ten minutes—a 101.8-fold speed-up. However, LabelStudio achieved higher accuracy (98\%) compared to Spacewalker (91\%), likely because Spacewalker is more complex. Labelled were more uniformly distributed among classes in LabelStudio, whereas users focused on label clusters in Spacewalker. Similar trends appeared in the text task, with LabelStudio achieving 82\% accuracy for 91 samples and Spacewalker achieving 72\% for 16,886.8 samples, reflecting a 185.6-fold speed-up at a 10\% accuracy trade-off.
Fig. \ref{fig:nasatlx_sus} (right) illustrates that users found the annotation in Spacewalker to be less time-consuming and requiring less effort compared to LabelStudio. However, users reported higher mental and physical demands for Spacewalker. While low mental and physical demands are desirable, extremely low scores might indicate boredom \cite{weinberg2016work}. The most striking difference was in the levels of frustration, with users reporting significantly less frustration when using Spacewalker. In the SUS questionnaires, users indicated a stronger preference for frequent use of Spacewalker over LabelStudio, consistent with NASA-TLX results. Spacewalker was seen as more intricate, which accounts for higher mental and physical demand scores and reduced confidence levels. Nonetheless, Spacewalker scored better for function integration. Both systems were rated low on being cumbersome and inconsistent, albeit with slight increases due to Spacewalker's novel controls. The perceived higher barrier to entry offers an explanation for this and is reinforced by a higher rating of needing support.

\subsection{Free Text User Remarks}
\label{sec:user_remarks}
Based on the verbatim comments, we summarize the impressions of the participants as follows:\\
\noindent\textbf{LabelStudio:} Participants found LabelStudio straightforward but uninspiring, with feedback emphasizing its intuitive design and ease of use. Positive remarks highlighted its clear interface and the confidence it provided in accurate labeling. However, criticisms focused on its slow speed and monotony, with some participants describing it as “boring” and “frustrating” after extended use.
\noindent\textbf{Spacewalker:} Users perceived Spacewalker as a more complex yet efficient tool, capable of handling large datasets quickly. While some reported initial challenges with the learning curve and configuration, they appreciated its advanced features and engaging elements, such as dynamic data visualization and gamification, which enhanced their experience and productivity.

% \noindent\textbf{LabelStudio} Participants found LabelStudio somewhat dull but acknowledged its ease of use due to its intuitive design. One participant described it as \textit{"intuitive, but pretty slow"} (P1) and as \textit{"ok, it's pretty boring and slow, but it works."} (P3). Main criticisms centered around the speed and repetitiveness of the tool: \textit{"Very frustrating, I can not use it for more than 10 mins"} (P14), \textit{"boring"} (P15). Positive remarks acknowledged the clarity of the interface and the confidence it instilled in their labelling: \textit{"going through each sample individually gives me confidence that I am labelling correctly."} (P16).

% \noindent\textbf{Spacewalker} The participants generally found Spacewalker to be a more complex but efficient tool: \textit{"Good, it’s quick, and you feel like you can learn the program and become even better and faster"} (P2), \textit{"annotate a large number of samples in a short time."} (P7). Some participants stated to have initial difficulties: \textit{"Good, but I should have taken more time at the beginning to better understand and optimally configure the different functions."} (P1), \textit{"it has a learning curve"} (P5), \textit{"it can be challenging to orient yourself in both views"} (P10). Despite these challenges, users appreciated various features that supported an engaging experience: "I valued being able to make adjustments to the plot" (P8), "I enjoyed the playful way to explore the data." (P19), "pretty cool, I really liked the gamification aspect" (P10).
\section{Discussion}
\label{sec:discussion}
This work shows that data exploration and annotation can be greatly improved with the right tools, especially for users without domain-specific expertise. In the study, participants accurately identified outliers during data integrity assessments without the need for programming skills, demonstrating that data exploration can become more intuitive and engaging. While t-SNE is popular in research, it falls short in scenarios requiring high interactivity and fast projections of new data points. In contrast, h-NNE performed better in these tasks. Additionally, 3D visualizations reduced point occlusion and improved user experience, with participants favoring tools like ours. This emphasizes the importance of developing user-friendly and interactive data analysis tools. In a data-driven world, we offer an alternative to traditional solutions for tasks ranging from business decisions to diagnostics.
In categorization tasks, Spacewalker achieved labelling rates more than 100 times faster than the baseline, with accuracy remaining within acceptable limits for most cases. CLIP was preferred because of its text querying capabilities, underscoring the need for seamless transitions between data types. Balancing speed, flexibility, and usability is crucial, as Spacewalker’s complexity can hinder quick familiarization. However, with interface improvements and user feedback, it has the potential to become a leading solution for large-scale data analysis. Future efforts should focus on refining the user experience, improving annotation accuracy, and expanding capabilities to meet the evolving needs of data analysts.
%Balancing speed, flexibility, and usability is essential for making Spacewalker a preferred solution in the rapidly advancing field of data visualization and annotation.

\section*{Acknowledgments}
This work was supported by DFG RTG 2535 and the Cancer Research Center Cologne Essen (CCCE).


\newpage

\bibliography{custom}

% \appendix

% \section{Appendix}
% \label{sec:appendix}
% \input{sections/appendix}

\end{document}